%% file: root.tex
\documentclass[letterpaper, 10 pt, conference]{ieeeconf}  

\IEEEoverridecommandlockouts                              

\usepackage{times}
\usepackage{geometry}

\usepackage{multicol}
\usepackage[bookmarks=true]{hyperref}
\usepackage{graphicx}
\usepackage{caption}
\usepackage{subcaption}
\usepackage{physics, amsmath}
\usepackage{xcolor}
\usepackage{graphicx}
\usepackage{siunitx}
\usepackage{multirow} 
\usepackage{amsmath}
\usepackage{amssymb}
\usepackage{algorithm}
\usepackage{algpseudocode}
\usepackage{booktabs}

\usepackage[
    style=ieee,
    sorting=none
]{biblatex}
\AtBeginBibliography{\small}

\title{\LARGE \bf
Action-Informed Estimation and Planning: Clearing Clutter on Staircases via Quadrupedal Pedipulation
}

\author{Prasanna Sriganesh, Barath Satheeshkumar, Anushree Sabnis and Matthew Travers
\thanks{All authors are from Robotics Institute, Carnegie Mellon University
        \scriptsize\texttt{\{pkettava, bsathees, asabnis, mtravers\}@andrew.cmu.edu}}%
\thanks{Supplementary Video - \scriptsize\texttt{\url{https://youtu.be/CTss0FFHwNU}}}
}

\begin{document}

\newgeometry{left = 54pt, top = 54pt, bottom = 54pt , right =54pt}

\maketitle
\thispagestyle{empty}
\pagestyle{empty}


\input{sections/abstract}

\IEEEpeerreviewmaketitle

\input{sections/introduction.tex}

\input{sections/related_work.tex}

\input{sections/technical_1.tex}

\input{sections/technical_2.tex}
\input{sections/experiments_results.tex}

\input{sections/conclusions.tex}

\printbibliography

\end{document}

%% file: sections/abstract.tex
\begin{abstract}

For robots to operate autonomously in densely cluttered environments, they must reason about and potentially physically interact with obstacles to clear a path. Safely clearing a path on challenging terrain, such as a cluttered staircase, requires controlled interaction. For example, a quadrupedal robot that pushes objects out of the way with one leg while maintaining a stable stance with its three other legs.  However, tightly coupled physical actions, such as one-legged pushing, create new constraints on the system that can be difficult to predict at design time. In this work, we present a new method that addresses one such constraint, wherein the object being pushed by a quadrupedal robot with one of its legs becomes occluded from the robot's sensors during manipulation. To address this challenge, we present a tightly coupled perception-action framework that enables the robot to perceive clutter, reason about feasible push paths, and execute the clearing maneuver. Our core contribution is an interaction-aware state estimation loop that uses proprioceptive feedback regarding foot contact and leg position to predict an object's displacement during the occlusion. This prediction guides the perception system to robustly re-detect the object after the interaction, closing the loop between action and sensing to enable accurate tracking even after partial pushes. Using this feedback allows the robot to learn from physical outcomes, reclassifying an object as immovable if a push fails due to it being too heavy. We present results of implementing our approach on a Boston Dynamics Spot robot that show our interaction-aware approach achieves higher task success rates and tracking accuracy in pushing objects on stairs compared to open-loop baselines.

\end{abstract}


\begin{keywords}
  Object Estimation and Tracking, Legged Robots, Planning and Control
\end{keywords}

%% file: sections/introduction.tex
\section{Introduction}

Successful real-world deployment of quadrupedal robots requires the ability to autonomously perceive, reason about, and physically navigate unpredictable environments. This requires them to go beyond simple avoidance and actively interact with obstacles, such as pushing objects that block their path. This capability becomes critical in complex terrains like a narrow, cluttered staircase, where a robot can neither treat items as static obstacles to be avoided nor step on them without compromising safety. Instead, the robot must reason about the clutter's physical properties to determine if a path can be safely cleared through interaction. For a quadruped, this action can be achieved by balancing on three legs while using the fourth to manipulate an object, an act termed ``pedipulation". This demands a tight integration of perception and action, as the robot must manage the physical push while dealing with inevitable perceptual challenges like occlusion. In this paper, we present a complete perception-to-action framework that addresses these challenges, enabling a quadruped robot to safely perceive, reason about, and clear movable clutter from a staircase.

\begin{figure} [t!]
    \begin{subfigure}[t]{\linewidth}
        \centering
        \includegraphics[width=0.7\linewidth]{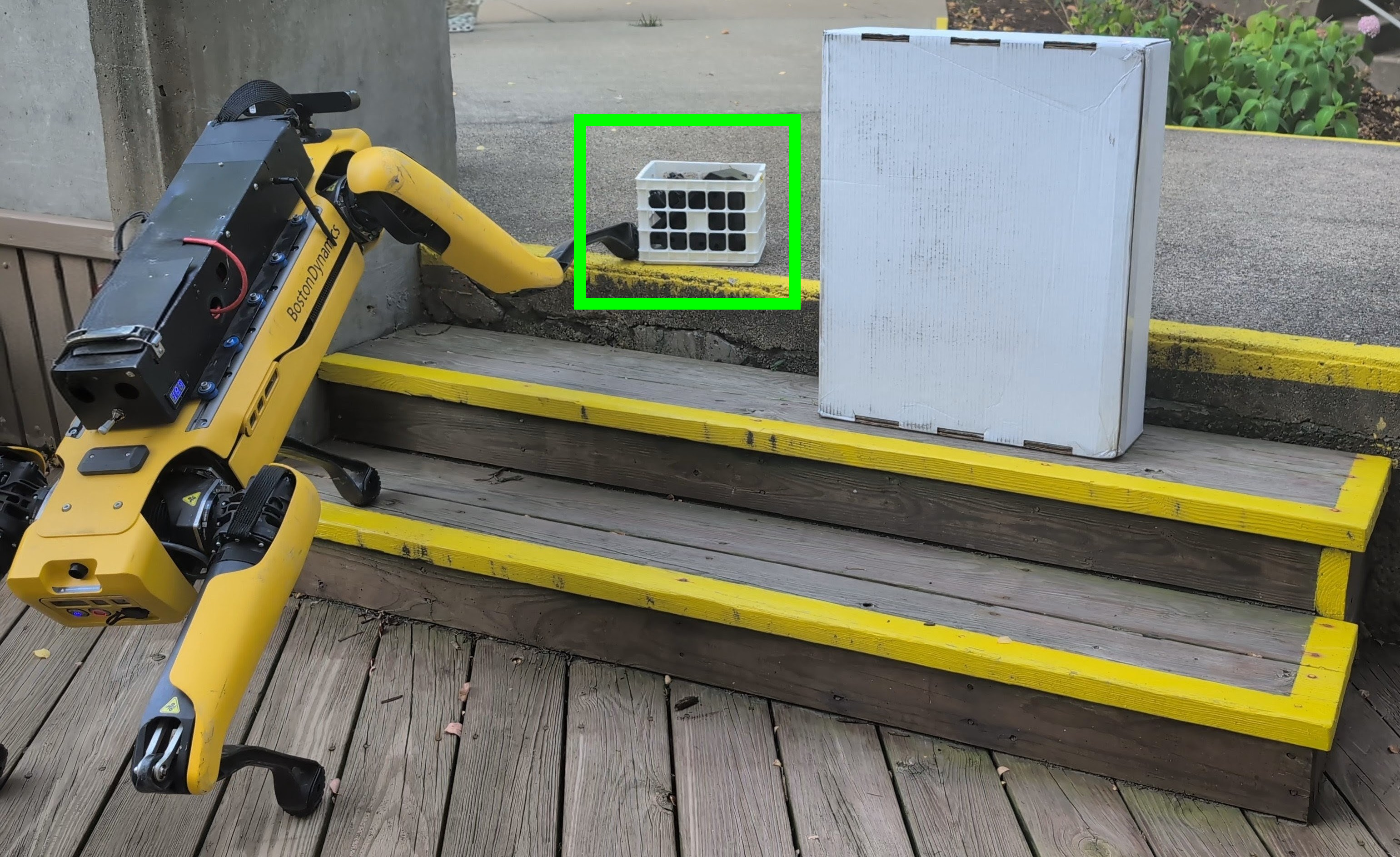}
    \end{subfigure}
    \begin{subfigure}[t]{\linewidth}
        \centering
        \includegraphics[width = 0.7\linewidth]{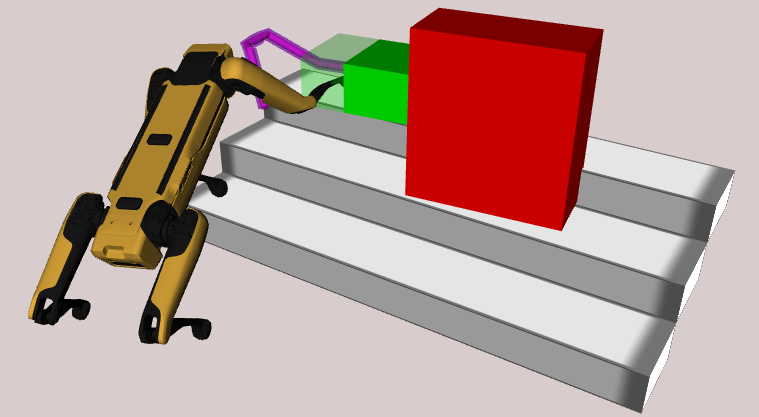}
    \end{subfigure}
    \caption{A spot robot pushing clutter to clear a path (Top). Visualization of our perceived world model: staircase shown with white markers, initial location of clutter is shown using translucent green marker, opaque green marker shows predicted location of object, and red marker shows immovable object. The foot path taken to push is shown in magenta (Bottom).}
    \label{fig:intro}
    \vspace{-1.5em}
\end{figure}

Many robotic pushing methods operate by decoupling perception from execution, an approach that is fundamentally limited in real-world scenarios. For example, during a push, the robot's motion to maintain balance while pushing causes the robot's body to tilt, shifting the camera's field-of-view away from the object on the stair. Additionally, the manipulating leg itself physically obstructs any remaining line of sight. Given these geometric constraints, achieving a consistently unobstructed view through additional sensors is infeasible. This loss of exteroception is consequential because push outcomes are not simple binary events. Factors like noisy state estimation and complex object shapes make partial pushes and foot slips common. Such unpredicted displacements can cause standard visual trackers to fail, leading the system to lose track of an object's state entirely.

This failure to track presents a new challenge--reasoning about the cause of a failed interaction. The robot must be able to distinguish whether an attempt failed because the foot missed the object or because it contacted an object too heavy to move. A robust system must be able to resolve this ambiguity in the failed push, and learn from it to intelligently update the robot's world model allowing it to reclassify an object as static or re-attempt the push.

To address these challenges, we introduce a perception framework that robustly segments clutter on staircases and maintains a low-dimensional world model of the environment. To overcome the challenge of visual occlusion within this model, it employs an interaction-aware state estimation loop. During a push, this loop uses proprioceptive feedback to predict an object's displacement, guiding the perception system to re-detect the object and update its state. This predict-correct cycle ensures robust tracking through complex interactions and updates key object properties, such as movability, within the world model. This state estimation is integrated into a hierarchical task execution framework. At the high level, a planner manages the robot's behavior by switching between navigation and pedipulation modes, while using contact feedback to learn from interactions to reclassify static objects, or retry pushes. It directs a low-level planner to compute safe foot trajectories for object pushing. Finally, a policy trained via reinforcement learning executes these trajectories, generating stable joint commands for the maneuver.

The main contributions of this work are:
\begin{itemize}
    \item A 3D perception framework that robustly segments and localizes clutter on staircases by leveraging geometric priors of the environment.
    \item An interaction-aware state estimation pipeline that robustly tracks an object's state through partial pushes and updates its movability by fusing visual measurements with proprioceptive contact feedback.
    \item A hierarchical planning and control architecture where a planning pipeline computes collision-free foot trajectories for pushing, and a learned pedipulation policy to executes these trajectories.
\end{itemize}
We validate our framework on a Boston Dynamics Spot robot across four cluttered staircases with objects of varying size, mass, and shape. The results demonstrate that our interaction-aware approach achieves higher tracking accuracy, which in turn leads to significantly improved task success rates compared to open-loop baselines.



%% file: sections/related_work.tex
\section{Related Work}

Our work sits at the intersection of quadrupedal manipulation and interaction-aware perception. We build upon research in both areas to create a complete system that can reason about and physically alter its environment.

\subsection{Manipulation with Legged Robots}

Research in quadrupedal manipulation has explored several directions. A common approach is to equip the robot with a dedicated manipulator arm \cite{Mittal2021ArticulatedOI, Chiu2022ACM}, which offers dexterity but often reduces the robot's payload capacity and introduces dynamic challenges to the locomotion controller. Another approach involves using the whole body of the robot to push large obstacles \cite{jeon2023learning, Shi2020CircusAA}. While effective for clearing paths, this method lacks the precision needed for more controlled interactions.


Several recent works have explored ``pedipulation" particularly for precise tasks like opening doors or pressing buttons \cite{arm2024pedipulate, Cheng2023LegsAM}. These methods primarily address the low-level control challenge, employing techniques like reinforcement learning (RL) to track end-effector target position in a whole-body fashion. Recent advancements have integrated local obstacle avoidance into a single policy \cite{stolle2024perceptive} or used hybrid RL and Model Predictive Control (MPC) for multi-limb coordination~\cite{zhu2025versatile}. While these policies demonstrate impressive dynamic control, they typically assume that the state of the target object is known and reliably tracked by an upstream perception system. Consequently, the challenge of closing the perception-action loop to verify task success or handle unexpected outcomes is generally considered outside the scope of these controllers.

The work most similar to ours in its aim to create an integrated system is that of Lu et al. \cite{lu2020autonomous_hexapod}, which presents a hexapod robot capable of perceiving and pushing objects. However, their approach relies on the inherent static stability of a hexapod, which allows for quasi-static manipulation on flat ground. The authors note that the manipulating leg occludes the camera's view, requiring the push to be executed open-loop using only proprioceptive feedback to follow a predefined Bézier curve trajectory. While they include a safety stop to prevent motor overloads during heavy object pushing, the system lacks a mechanism to verify the outcome of the push, track partial displacements, or update the object's state in a persistent world model. In contrast, our work must address the dynamic stability challenges of a quadruped on a staircase and introduces an interaction-aware perception loop to explicitly handle the uncertainty of these interactions.


\subsection{Perception for Dynamic Objects}

A related line of research focuses on the perception and tracking of dynamic objects. Many learning-based methods are trained to segment and track specific object categories, such as cars or pedestrians, and use filtering approaches to track them \cite{khatri2024trackflow, xu2023real}. Another common technique is scene flow estimation, which computes the velocity of 3D points in the environment \cite{chen2021moving, sun2022efficient}. More recently, Vision-Language Models have also been explored for this task \cite{sapkota2025review}. However, these methods either have limited adaptability to unseen data, require continuous visibility of the object, or are too computationally intensive for real-time use on mobile robots.

\begin{figure*}[!t]
    \centering
    \includegraphics[width = 0.8\linewidth]{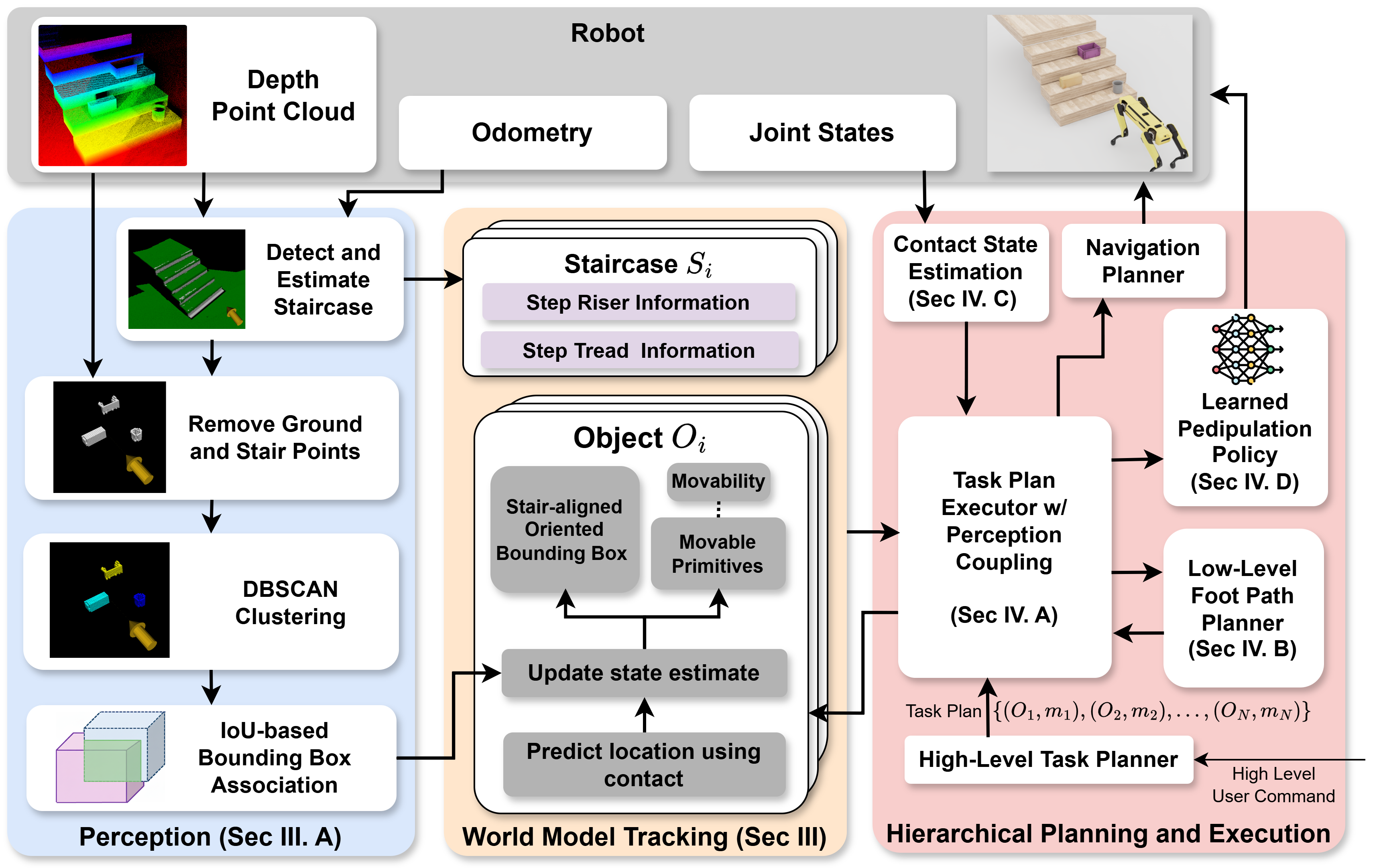}
    \caption{Overview of our proposed framework for perceiving, tracking and pushing objects on cluttered staircases}
    \label{fig:full_diag}
    \vspace{-1.5em}
\end{figure*}

A more general approach involves using a map of the static world to detect dynamic changes. This idea was first explored in \cite{modayil2008initial}. Dynablox \cite{schmid2023dynablox} proposed a method to segment dynamic objects in the scene by estimating high-confidence free space areas in the map. However, this requires careful modeling of the state estimation and mapping drift and does not perform temporal tracking of the object. Furthermore, these methods are not designed for scenarios where the robot itself is the cause of the object's motion. Our work is fundamentally concerned with this setting, where the system must predict the physical consequences of its own actions and use that prediction track an object under occlusion. This requires the perception system to utilize feedback from the planning and control layers.

%% file: sections/technical_1.tex
\section{Clutter Movability Estimation}

\label{sec:perception}

To enable interaction, the robot must first build a persistent and accurate model of its environment from raw sensor data. Our perception pipeline, illustrated in the left column of Fig. \ref{fig:full_diag}, is designed to process incoming point clouds to detect staircases, segment potential clutter, and maintain a consistent track of each object's state through physical interaction. This process begins with detecting and modeling the static environment, which then allows us to isolate and reason about the dynamic objects within it.

\subsection{World Modeling and Clutter Initialization}
The perception pipeline populates and maintains a low-dimensional world model, $\mathcal{W}$, which serves as the central state representation for the planning framework. This model consists of three primary elements: a set of staircases $\{S_i\}$, a set of objects $\{O_i\}$, the ground plane $G$. Each staircase $S_i$ is represented by its geometric properties.

The representation for each object $O_i$ is central to our contribution. Each object is assigned a unique ID for tracking and stores its associated point cloud cluster, $P_i$. Its physical state is modeled as an oriented bounding box (OBB), defined by its center position and dimensions. Critically, to create a stable and structured representation suitable for manipulation planning, we constrain the orientation of the bounding box to be aligned with the principal direction of the stair step on which it rests. The object's movability state is discrete: it is either classified as \textit{static} or is associated with a set of \textit{movable motion primitives} $\{m_k\}$. These primitives, described in Sec. \ref{sec:primitive_generation}, define stable paths along which the object can be pushed.

\begin{figure}[b!]
    \centering
    \vspace{-1.0em}
    \begin{subfigure}[t]{0.26\linewidth}
        \centering
        \includegraphics[width = \linewidth]{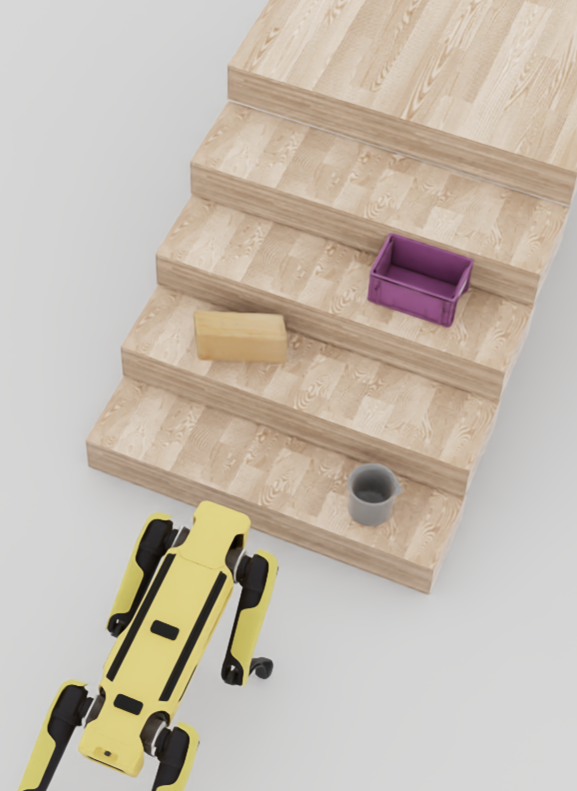}
        \caption{}
        \label{fig:sim_example}
    \end{subfigure}
    \begin{subfigure}[t]{0.29\linewidth}
        \centering
        \includegraphics[width = \linewidth]{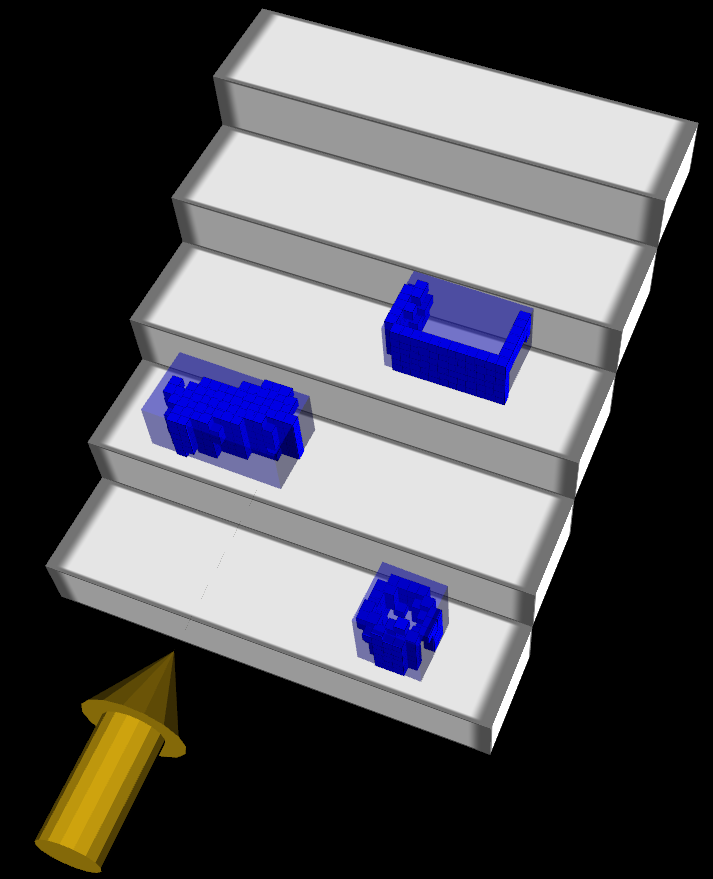}
        \caption{}
         \label{fig:obb_visualization}
    \end{subfigure}
    \begin{subfigure}[t]{0.3145\linewidth}
        \centering
        \includegraphics[width =\linewidth]{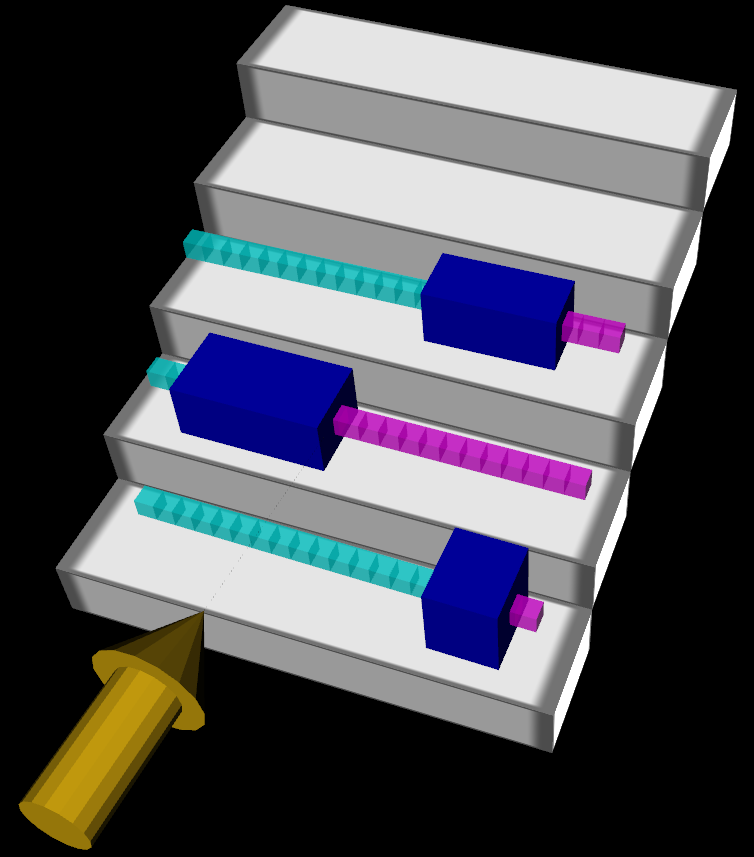}
        \caption{}
         \label{fig:primitive_visualization}
    \end{subfigure}
    \caption{Example of object detection and movable primitive generation. (a) Robot near a cluttered staircase (b) Estimated staircase highlighted by white marker, translucent blue marker denotes the stair-aligned oriented bounding box, and blue points represent the object point cloud (c) `Movable primitives' generated for each object for two directions shown with cyan and magenta paths}
    \label{fig:opening_results}
\end{figure}

The process of building this model begins with a registered point cloud from the robot's depth camera. First, we identify the traversable ground surfaces by leveraging a robust method for staircase estimation and ground segmentation~\cite{stairdetection}\cite{sriganesh2025bayesian}. This provides us with a segmented point cloud of all clutter-free surfaces, including the stair treads, stair risers and the initial ground plane. To isolate the clutter, we perform a subtraction: all points identified as ground or stair surfaces are removed from the initial point cloud. The remaining points are assumed to be objects. We then apply the DBSCAN clustering algorithm \cite{schubert2017dbscan} to these object points. For each resulting cluster, $C_i$, an initial OBB is fit, with its orientation constrained by the underlying stair geometry. The dimensions of this OBB are then checked against a predefined threshold based on the robot's capabilities. Clusters larger than this threshold are immediately classified as \textit{static}. If objects already exist in the current world model instance, these new clusters are passed to our tracking pipeline (Sec. \ref{sec:clutter_tracking}) for data association and state updates.Clusters that do not match an existing object are used to initialize a new instance $O_i$ in the world model. These new objects are assigned a `potentially movable' status. Fig. \ref{fig:obb_visualization} showcases the computed stair-aligned OBB for a simple cluttered staircase scenario.

\subsection{Movable Primitive Generation}
\label{sec:primitive_generation}

For each object classified as potentially movable, we generate a set of valid motion primitives, $\{m_k\}$. These `movable primitives' represent kinematically feasible paths to push an object from its current location to the edge of the staircase, thereby clearing a path for the robot. For the structured case of a staircase, we generate primitives in the two directions perpendicular to the main axis of the stair step (i.e., push left and push right). 

A primitive path is generated by simulating a sequence of small, incremental translations of the object's OBB. The simulation effectively traces the object's trajectory under a push action originating at the current object's position.
The path begins at the center of the face of the object's OBB that's farthest from the direction of its intended motion (e.g., left-most face for pushing right). The path terminates when the object reaches the boundary of the traversable area. This boundary is defined as the first point where one of two conditions is met:
\begin{enumerate}
    \item \textbf{Collision-freedom:} The OBB, at each translated pose, must not intersect with any static elements of the world model, such as the staircase structure or any other objects in the scene. A collision terminates the primitive path.
    \item \textbf{Support Stability:} The object must remain stably supported. The primitive is terminated if any of the four vertices of the OBB's bottom face lose contact with a known navigable surface (i.e., stair treads or the ground plane). This correctly identifies the edge of an open-sided step.
\end{enumerate}
The resulting set of collision-free and stable primitives is stored with the object $O_i$ in the world model, providing the planner with a set of guaranteed-safe manipulation actions to clear a path.  Fig. \ref{fig:primitive_visualization} showcases the computed movable primitives for a simple cluttered staircase scenario.

\subsection{Interaction-Aware Clutter Tracking}
\label{sec:clutter_tracking}

A core challenge in interaction-aware perception is maintaining a consistent world model, especially after an object's state is changed by the robot's own actions. Our framework employs a tracking approach that combines standard visual data association with a novel proprioceptive tracking method to handle periods of visual occlusion during a push.

\textbf{Visual Data Association:} In each perception cycle, new object clusters $\{C_j\}$ are matched to existing objects $\{O_i\}$ in the world model. The target for this matching depends on whether an interaction has just occurred. In the absence of interaction, the target is the object's last known OBB. However, after a push, the target is the \textit{predicted OBB} generated via proprioceptive feedback. A match is found by computing the Intersection over Union (IoU) between the new cluster's OBB and the target OBB. If successful, the new point cloud $P_j$ is merged with the existing object's cloud $P_i$ using the Iterative Closest Point (ICP) algorithm to refine the object's geometry and compute the new OBB.

\textbf{Proprioceptive State Prediction:} During a push, the robot's camera loses sight of the object, making visual tracking impossible. A purely open-loop prediction, assuming the object moves to the end of its motion primitive, would fail to account for partial pushes. To overcome this, we use proprioceptive feedback to generate a more accurate prediction. When the executor signals contact, the system assumes the object's OBB moves rigidly with the foot's end-effector. To prevent prediction drift from foot slippage, the system continuously finds the closest point on the commanded motion primitive to the foot's current position. The object's predicted OBB is then updated based on the corresponding translation along this primitive, not the raw foot motion.

This contact-aware prediction is the key to tracking through interaction. When the push is complete and vision is restored, the new visual cluster appears at the object's true, potentially partially pushed location. The visual data association then compares this new cluster not with the object's original OBB, but with the much more accurate proprioceptively-predicted OBB. Because this prediction correctly accounts for partial pushes, the IoU is high, leading to a successful match and a consistent state update. This predict-correct cycle allows the system to robustly track an object's state even after unpredictable interactions that would cause vision-only trackers to fail.

%% file: sections/technical_2.tex
\section{Hierarchical Planning and Execution}

To clear clutter from a staircase, the robot must execute a sequence of complex behaviors that tightly couple navigation, manipulation, and perception. We propose a hierarchical framework (Fig. \ref{fig:full_diag}) that decomposes this task into three manageable layers: a high-level task executor that orchestrates the overall mission, a low-level path planner that computes collision-free trajectories for the foot, and a learned pedipulation policy that translates these trajectories into dynamic, stable motor commands. This architecture allows the system to reason about high-level goals while reacting to low-level physical feedback.

\subsection{Task Plan Executor with Perception  Coupling}

The core of our planning framework is the Task Plan Executor, which receives a high-level task plan and orchestrates the low-level actions required to complete it. A task plan is defined as a sequence of manipulation primitives $\{ (O_1, m_1), (O_2, m_2), \dots, (O_N, m_N) \}$, where each pair specifies a target object $O_i$ and a desired motion primitive $m_j$. While this plan can be generated by various methods, such as a navigation among movable obstacles (NAMO) planner~\cite{namo1}~\cite{namo2} or even vision-language models (VLMs)~\cite{vlm_planning}, the specifics of the high-level planner are outside the scope of this work. Our focus is on the robust execution of a given plan.

The executor processes one manipulation task at a time, as detailed in Algorithm \ref{alg:executor}. For each task, it first invokes a navigation planner to align the robot's body with the target object, creating a suitable workspace for manipulation (Line~\ref{line:start_align}). This planner commands velocity commands to the robot's walking controller to reach the target pose. Once aligned, the executor calls the low-level foot path planner to generate a collision-free trajectory, $\mathcal{T}_{\text{foot}}$, for the selected leg (Line~\ref{line:plan_foot}). If a valid path is found, the executor passes each waypoint sequentially to the pedipulation policy for execution (Line~\ref{line:exec_loop}).

The executor's tight coupling with the perception system is critical during physical interaction. While the policy tracks the foot trajectory, the executor continuously monitors the leg's contact state (Line \ref{line:get_contact}). Upon making contact, it signals the world model to begin its proprioceptive state tracking (Line \ref{line:signal_contact}). Concurrently, it monitors for interaction failures. If the foot is in contact but its position remains unchanged for a 5-second window, the system infers that the object is immovable (Line \ref{line:check_stall}). This triggers an update to the world model, reclassifying the object's movability property to \textit{static}, and the current push is aborted (Lines \ref{line:set_static}-\ref{line:return_static}). After the entire push attempt is complete and the foot has returned to a safe resting pose (Lines \ref{line:plan_return}-\ref{line:exec_return}), the executor evaluates the outcome. It compares the object's new state (which has been updated by a new visual measurement, using the initial proprioceptive prediction), with the expected end state from the motion primitive. If a significant discrepancy exists (i.e., a partial push), the executor retries the entire manipulation task for that object from the beginning (Lines \ref{line:check_partial}-\ref{line:goto}). Upon successfully completing the sequence of manipulation tasks, the executor transitions back to a navigation mode, commanding the robot to ascend the now-cleared staircase to complete the high-level mission.

\begin{algorithm} [t!]
\caption{Task Plan Executor Logic for a Single Push}
\label{alg:executor}
\begin{algorithmic}[1]
\Require Manipulation Task $(O_i, m_j)$ 
\Require World Model $\mathcal{W}$
\Ensure Updated World Model $\mathcal{W}'$
\State \texttt{AlignRobotToObject($O_i$)} \label{line:start_align} 
\State $leg_{\texttt{id}} \gets$ \texttt{SelectManipulationLeg($O_i, m_j$)} 
\State $\mathcal{T}_{\texttt{foot}} \gets$ \texttt{PlanFootTrajectory($O_i, m_j ,\mathcal{W}$)} \label{line:plan_foot}
\For{each foot waypoint $p_{\texttt{goal}}$ in $\mathcal{T}_{\texttt{foot}}$} \label{line:exec_loop}
    \State \texttt{ExecuteWaypoint($p_{\text{goal}}, leg_{\text{id}}$)}
    \State $C \gets$ \texttt{GetContactState($leg_{\texttt{id}}$)} \label{line:get_contact}
    \If{$C$ is \texttt{IN\_CONTACT}}
        \State \texttt{UpdateWorldModelContact}($\mathcal{W}, O_i, \texttt{True}$)\label{line:signal_contact} 
        \State $p_{\texttt{foot}} \gets$ \texttt{GetFootPosition($leg_{\texttt{id}}$)}
        \If{\texttt{IsPositionStalled}($p_{\texttt{foot}}$, $5\;sec$)} \label{line:check_stall}
            \State \texttt{SetObjectMovability($O_i, \texttt{STATIC}$)} \label{line:set_static}
            \State \texttt{AbortPush}()
            \State \Return $\mathcal{W}$ \label{line:return_static}
        \EndIf
    \EndIf
\EndFor
\State $\mathcal{T}_{\text{return}} \gets$ \texttt{PlanReturnTrajectory}($\mathcal{W}$) \label{line:plan_return}
\For{each foot waypoint $p_{\texttt{return}}$ in $\mathcal{T}_{\texttt{return}}$} \label{line:exec_return_loop}
    \State \texttt{ExecuteWaypoint}($p_{\texttt{return}}, leg_{\texttt{id}}$)
\EndFor \label{line:exec_return}
\State $p_{\text{obj\_actual}} \gets \mathcal{W}$\texttt{.GetObjectPosition}($O_i$) \label{line:check_partial}
\State $p_{\text{obj\_expected}} \gets m_j$\texttt{.GetExpectedEndPosition}()
\If{\texttt{distance}$(p_{\text{obj\_actual}}, p_{\text{obj\_expected}}) > \texttt{THRESH}$}
    \State \textbf{goto} \ref{line:start_align} \Comment{Partial push detected, retry} \label{line:goto}
\EndIf
\State \Return $\mathcal{W}$
\end{algorithmic}
\end{algorithm}
\subsection{Low-Level Foot Path Planning}

The low-level planner is responsible for generating collision-free 3D paths for the manipulating leg's end-effector. It uses a standard A* search algorithm operating on a 3D voxel grid representation of the world model, which includes both the staircase geometry and all perceived objects. Given a target motion primitive from the executor, the planner first computes an \textit{approach path} from the foot's current resting position to the primitive's starting pose. It then concatenates this approach path with the push path defined by the motion primitive itself to form the complete manipulation trajectory, $\mathcal{T}_{\text{foot}}$. After the push is complete, it similarly computes a \textit{return path} to guide the foot back to a safe resting stance.

\subsection{Contact State Estimation}
\label{sec:contact}
To enable the executor's interaction-aware logic, the system must reliably detect when the manipulating leg makes contact with an object. We achieve this by monitoring motor torques for unexpected efforts. We compute a residual torque, $\tau_{\text{residual}}$, by taking the difference between the measured actuator torques, $\tau_{\text{measured}}$, and the torques predicted by the full rigid-body dynamics of the robot, $\tau_{\text{model}}$, as:
\begin{align}
    \tau_{\text{residual}} &= \tau_{\text{measured}} - \tau_{\text{model}} \nonumber \\
    \tau_{\text{residual}} &= \tau_{\text{measured}} - (M(q)\ddot{q} + C(q, \dot{q})\dot{q} + G(q))
    \label{eq:residual_torque}
\end{align}
where $q$, $\dot{q}$, and $\ddot{q}$ are the joint positions, velocities, and accelerations, $M(q)$ is the mass matrix, $C(q, \dot{q})$ accounts for Coriolis and centrifugal forces, and $G(q)$ is the gravity vector. 
During a push maneuver, the robot's supporting legs remain stationary and its body motion is quasi-static. While the dynamics model computes expected torques for the entire robot, these assumptions allow us to reliably isolate the source of any external force to the manipulating leg. To detect contact, we therefore compute the L2 norm of the $\tau_{\text{residual}}$ vector for only the pedipulating leg's joints. If this magnitude exceeds a threshold for a sustained period, we confirm that contact has been made.

\subsection{Learned Pedipulation Policy}

The execution of the foot trajectories is managed by a learned reinforcement learning (RL) policy. Our policy builds upon the framework presented in \cite{stolle2024perceptive}, which generates low-level joint commands to maintain stability while tracking end-effector targets. We adopt the base proprioceptive observations from this work and augment them with our leg selection flags, as detailed in Table \ref{tab:rl_obs}. Our primary contributions to the policy are threefold. 

First, to allow the Task Plan Executor to seamlessly switch between modes, we augment the policy's command input with binary flags, $[a_L, a_R]$, to specify the manipulating leg (front-left or front-right). This is coupled with a modified reward structure: the primary task reward encourages the active leg to track its target foot position, while an additional reward for stable standing is used when no manipulation is commanded ($a_L=a_R=0$). This standing reward encourages the policy to keep all four feet in contact with the ground with minimal body motion. Normalization rewards penalizing high joint torques and velocities follow the structure in \cite{stolle2024perceptive} to ensure safe execution.

Second, to improve training efficiency and promote generalization, we treat object interaction as an external disturbance to be corrected. During training, randomized external forces are applied to the end-effector of the manipulating leg, encouraging the policy to learn a robust response without simulating expensive rigid-body contacts. Finally, we also include staircases in the training environments to ensure the policy is adept at handling the specific terrain of our target application. We refer the reader to \cite{stolle2024perceptive} for complete details on the training curriculum and reward terms.

\begin{table}[h]
\centering
\begin{tabular}{@{}llc@{}}
\toprule
\textbf{Observation} & \textbf{Symbol} \\ \midrule
Base Linear and Angular Velocity & ${}^\mathcal{B} v_B$ ${}^\mathcal{B} \omega_B$ \\
Projected Gravity Vector & ${}^\mathcal{B} g$ \\
Joint Positions and Joint Velocity& $q_j$, $\dot{q}_j$  \\
Foot Position Command & ${}^\mathcal{B} p_f$ \\
\textbf{Manipulation Leg Flags} & $[a_L, a_R]$ \\
Last Actions & $a_{t-1}$ \\
Height Scan & $[h_{1,1}, h_{1,2}, \dots h_{l, k}]$ \\
\bottomrule
\end{tabular}
\caption{Policy observations. The superscript $\mathcal{B}$ denotes that vectors are expressed in the robot's body frame. Terms in bold indicate our additions to \cite{stolle2024perceptive}}
\label{tab:rl_obs}
\vspace{-1.5em}
\end{table}

%% file: sections/experiments_results.tex
\section{Experiments and Results}
\subsection{Experiments}

\textbf{System Implementation:} Our framework was deployed on a Boston Dynamics Spot robot equipped with an NVIDIA Jetson AGX Orin for onboard computation. Perception was performed using a front-mounted ZED 2i Stereo Camera to generate point clouds. The robot's state estimation for localization was provided by the proprietary onboard odometry system. For the low-level controllers, the pedipulation policy was trained using Proximal Policy Optimization (PPO) in Isaac Lab over three days on a single NVIDIA RTX 4080 Super GPU. The navigation tasks, such as aligning the robot with an object, were handled by the local planner provided with the Spot SDK.

\textbf{Experimental Validation:} We evaluated our system across 4 different real-world staircases using 4 distinct objects of varying size, shape, and mass. Some of these objects' mass was increased to test the movability update scenarios. To assess overall performance, we measured the task success rate over 40 trials. A trial is considered a success if either (1) the object is successfully pushed to the edge of the staircase (even with retries in case of partial pushes) and its final state is correctly tracked, or (2) a heavy object is correctly identified as immovable and its state is updated to \textit{static}. Conversely, a trial is deemed a failure if the tracking pipeline loses the object's identity after a push, as this prevents the system from intelligently reattempting the task based on the object's new location. To specifically validate our contact-based tracking pipeline, we analyzed the accuracy of its state prediction. After each push maneuver, we measured the mean prediction error, defined as the difference between the object's proprioceptively predicted pose and its actual pose as determined by the subsequent visual update. This error was compared against that of an open-loop baseline. This baseline uses the same planning and control stack but predicts the object's new position to be at the end of the planned motion primitive, without using any contact feedback for correction.

\begin{figure}[t!]
    \centering
    \begin{subfigure}[t]{0.262\linewidth}
        \centering
        \includegraphics[width = \linewidth]{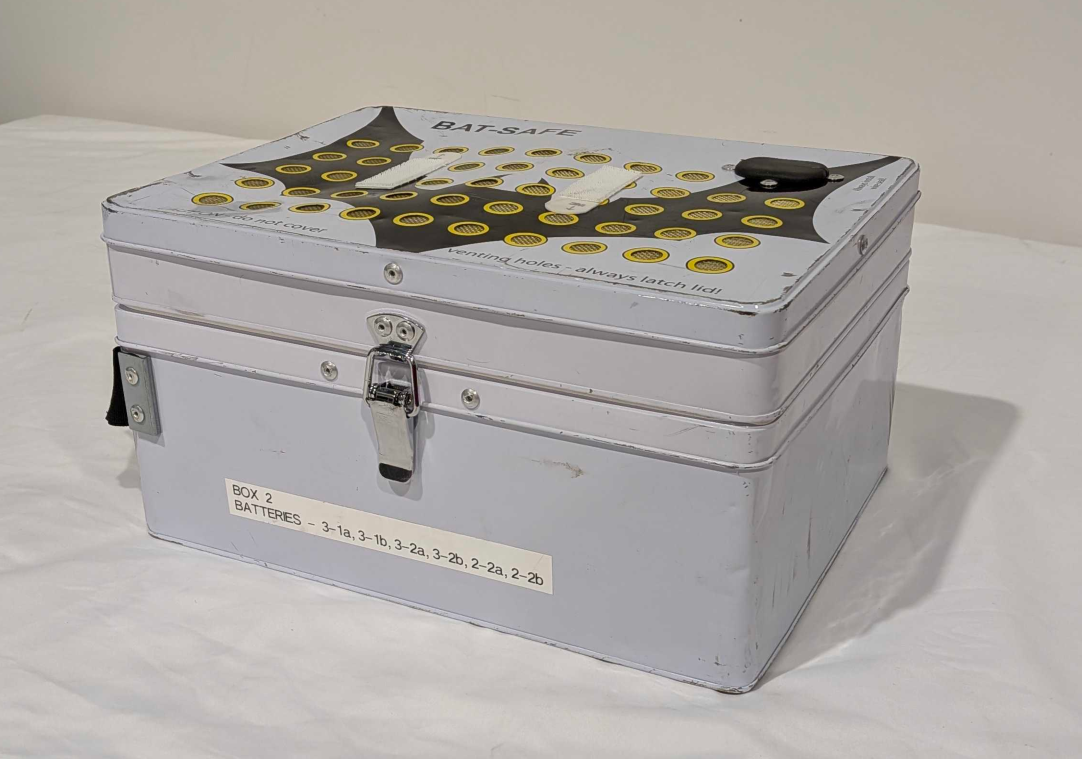}
        \caption{}
    \end{subfigure}
    \begin{subfigure}[t]{0.135\linewidth}
        \centering
        \includegraphics[width = \linewidth]{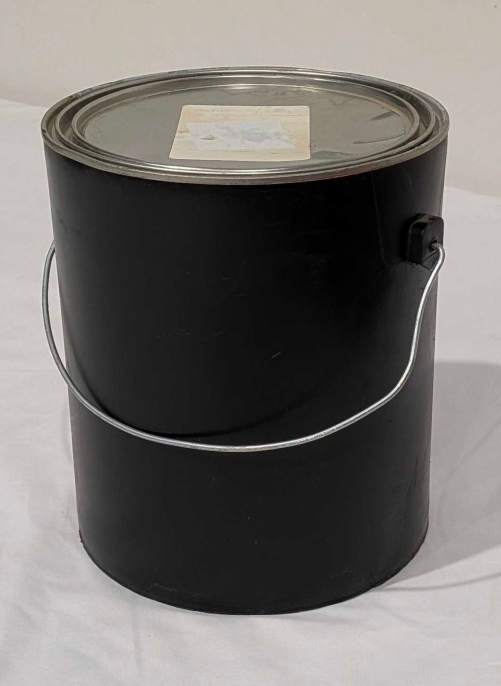}
        \caption{}
    \end{subfigure}
    \begin{subfigure}[t]{0.24\linewidth}
        \centering
        \includegraphics[width =\linewidth]{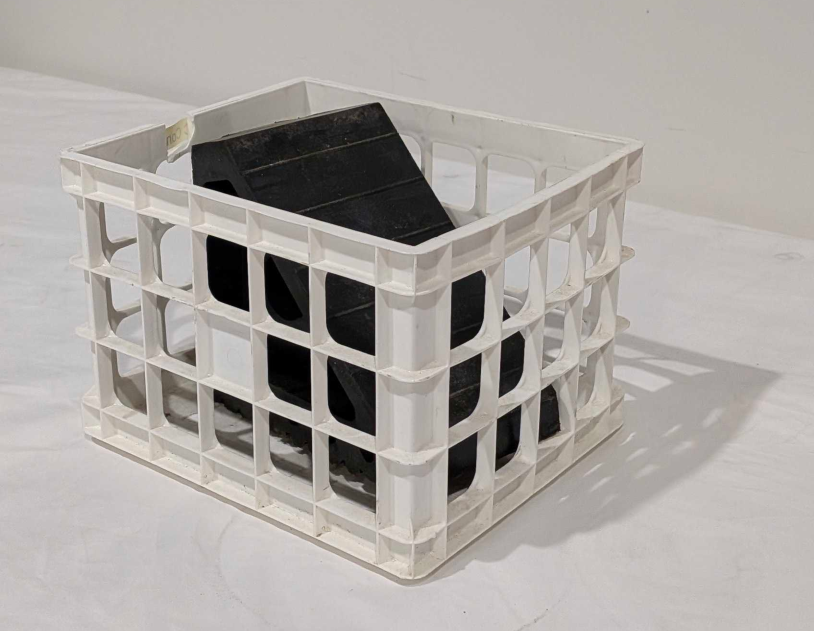}
        \caption{}
    \end{subfigure}
    \begin{subfigure}[t]{0.3\linewidth}
        \centering
        \includegraphics[width =\linewidth]{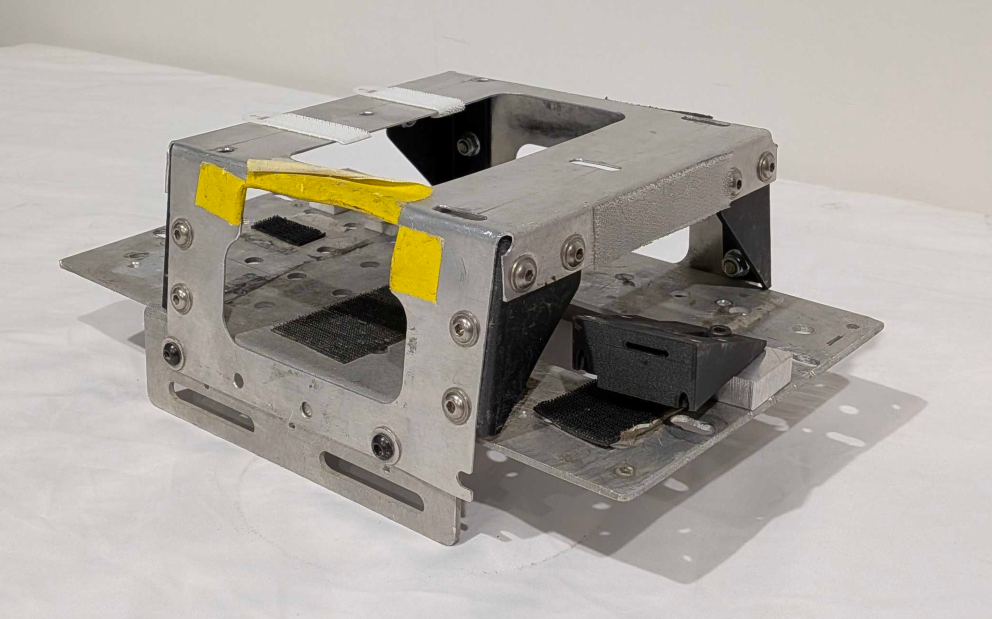}
        \caption{}
    \end{subfigure}
    \caption{Different objects used to validate our proposed framework. (a) Box (b) Paint can (c) Weighted crate (d) Metal frame}
    \label{fig:objects}
    \vspace{-1.5em}
\end{figure}

\subsection{Results}
\begin{figure}[!b]
    \centering
    \vspace{-1.0em}
    \includegraphics[width=0.9\linewidth]{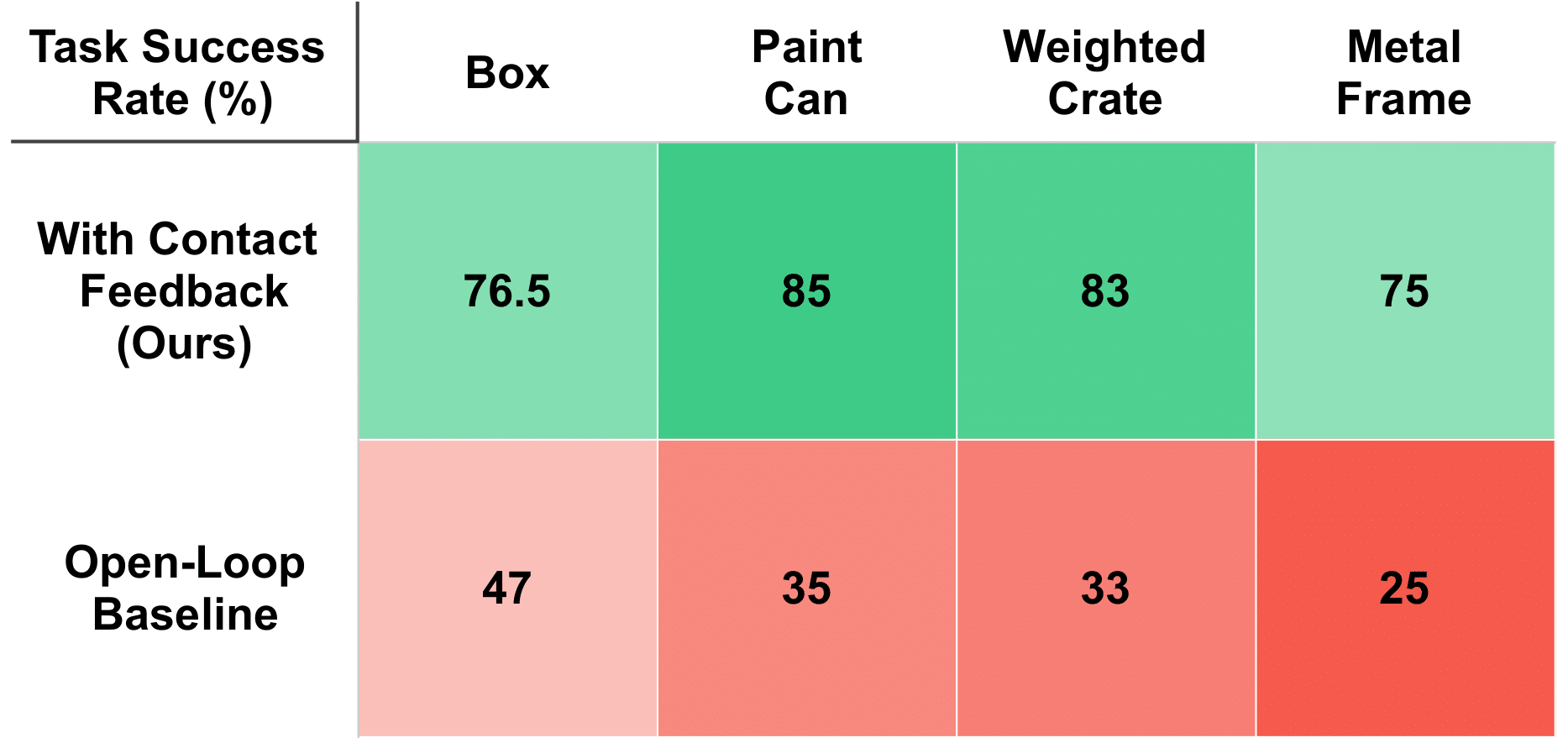}
    \caption{Success Rate comparison of push tasks for different objects}
    \label{fig:success_rate}
\end{figure}

\begin{figure*}[!t]
    \centering
    \begin{subfigure}[t]{.32\linewidth}
        \centering
        \includegraphics[width = \linewidth]{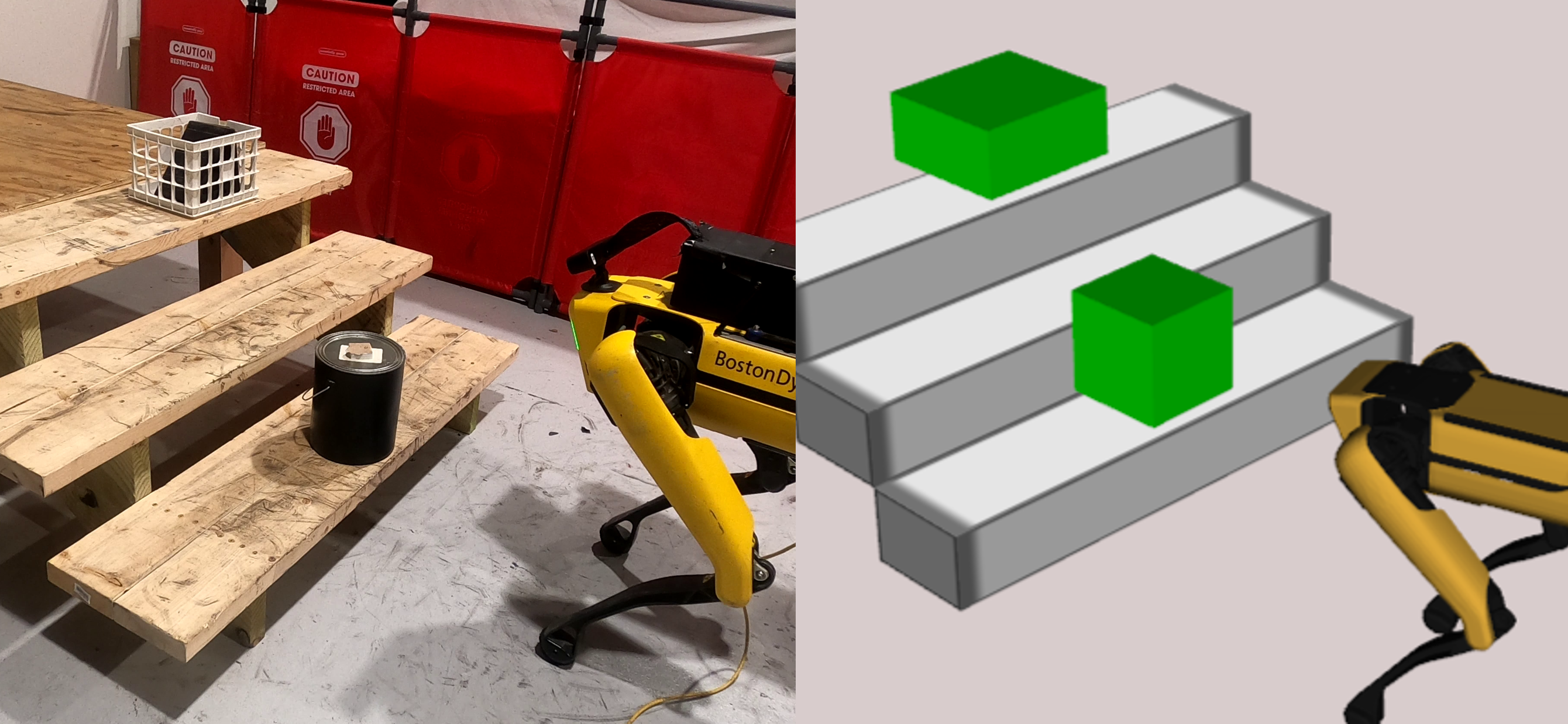}
    \end{subfigure}
    \begin{subfigure}[t]{.32\linewidth}
        \centering
         \includegraphics[width = \linewidth]{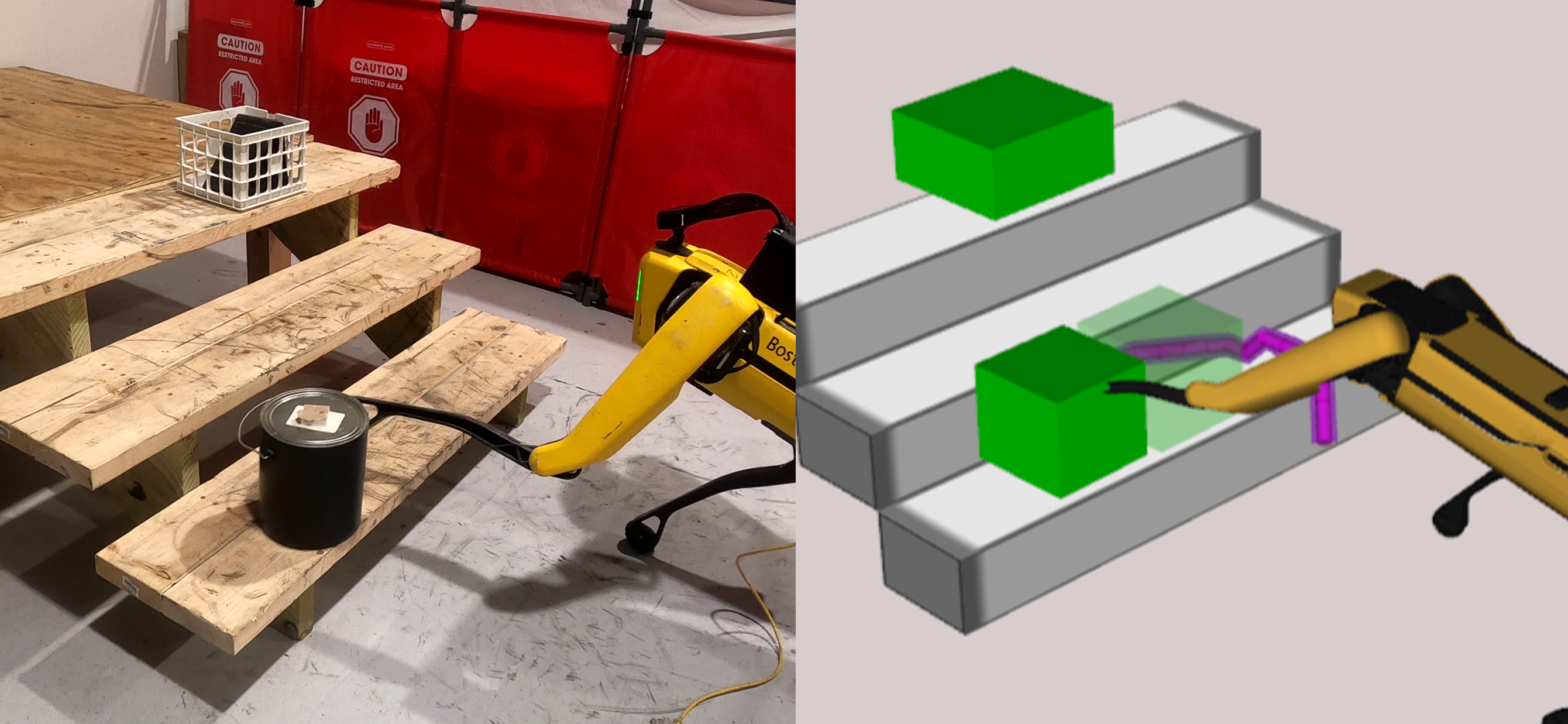}
    \end{subfigure}
    \begin{subfigure}[t]{.32\linewidth}
        \centering
         \includegraphics[width = \linewidth]{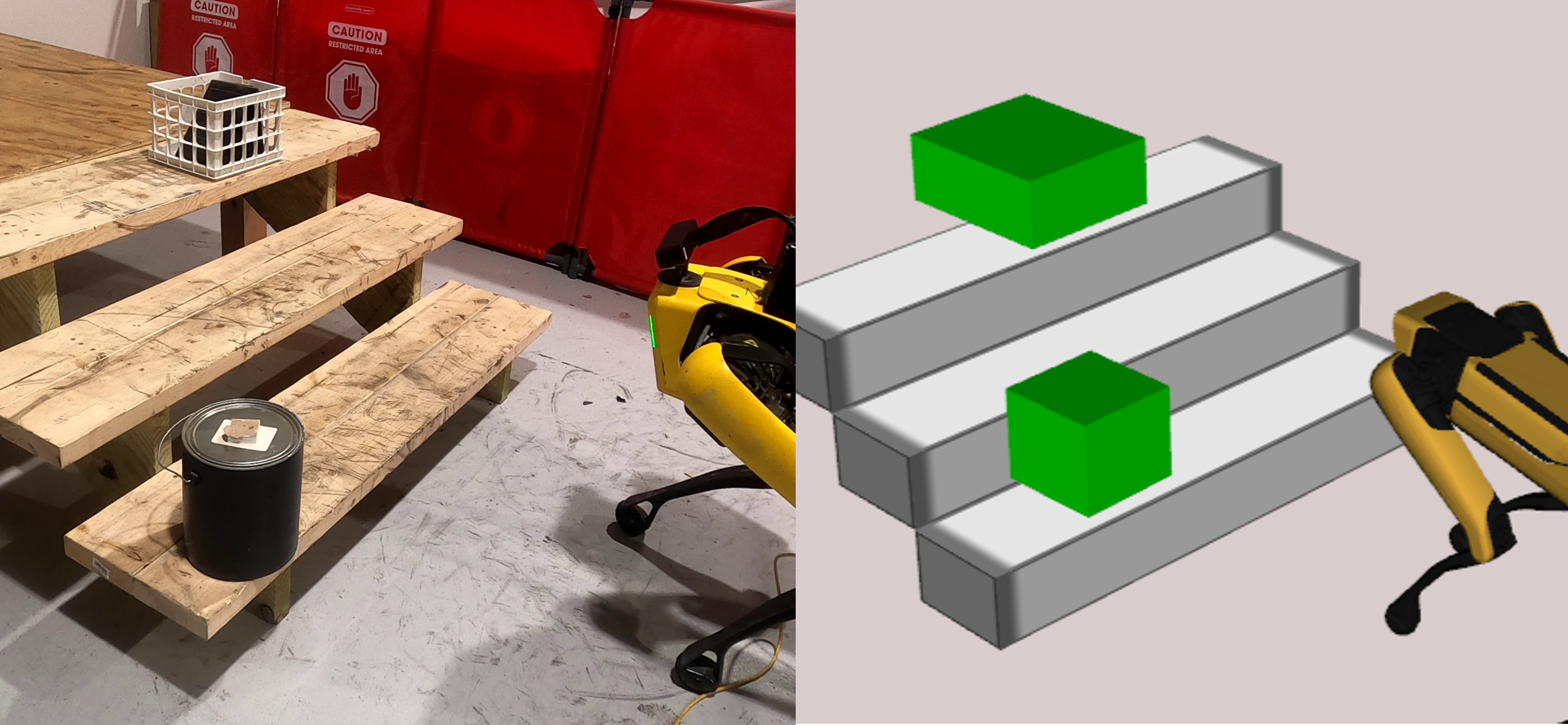}
    \end{subfigure}
    \begin{subfigure}[t]{.32\linewidth}
        \centering
        \includegraphics[width = \linewidth]{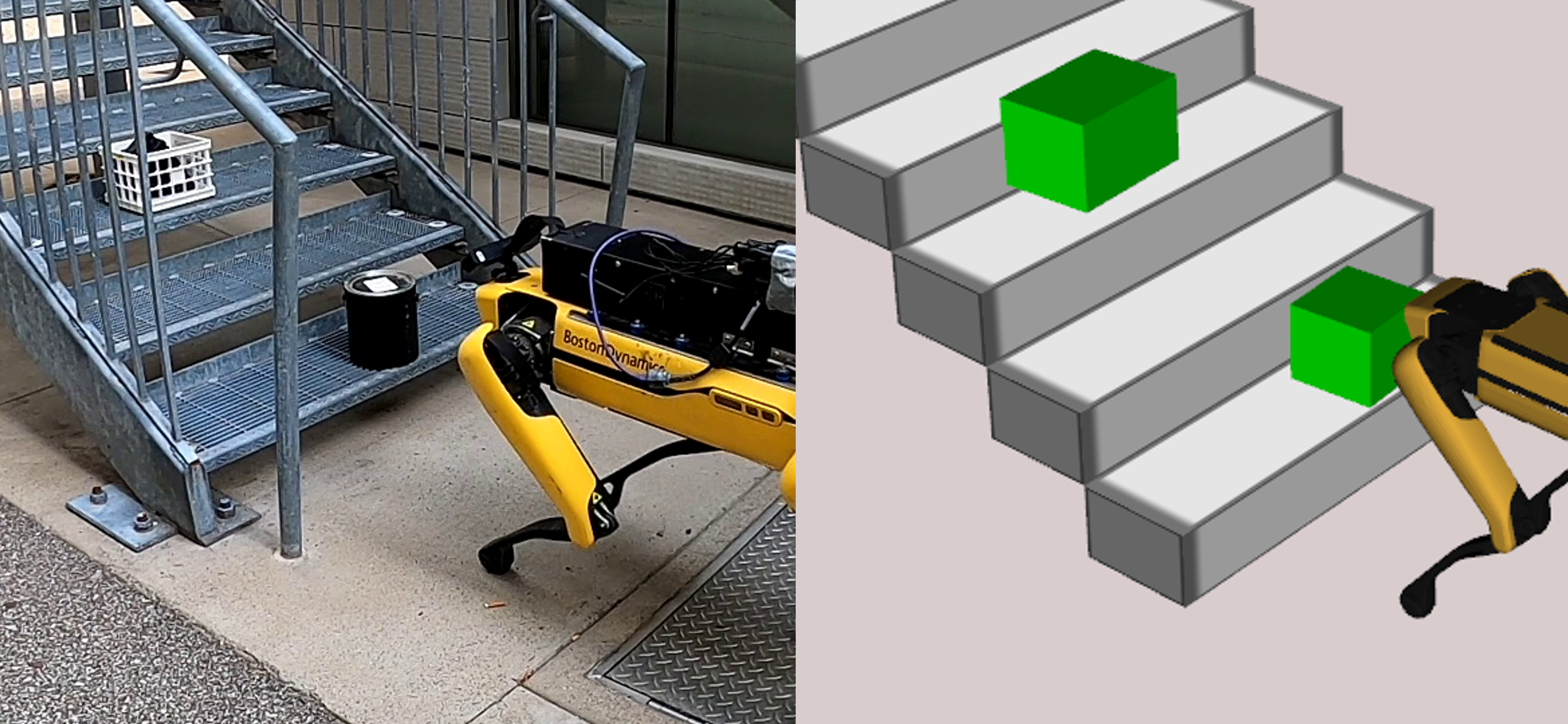}
    \end{subfigure}
    \begin{subfigure}[t]{.32\linewidth}
        \centering
         \includegraphics[width = \linewidth]{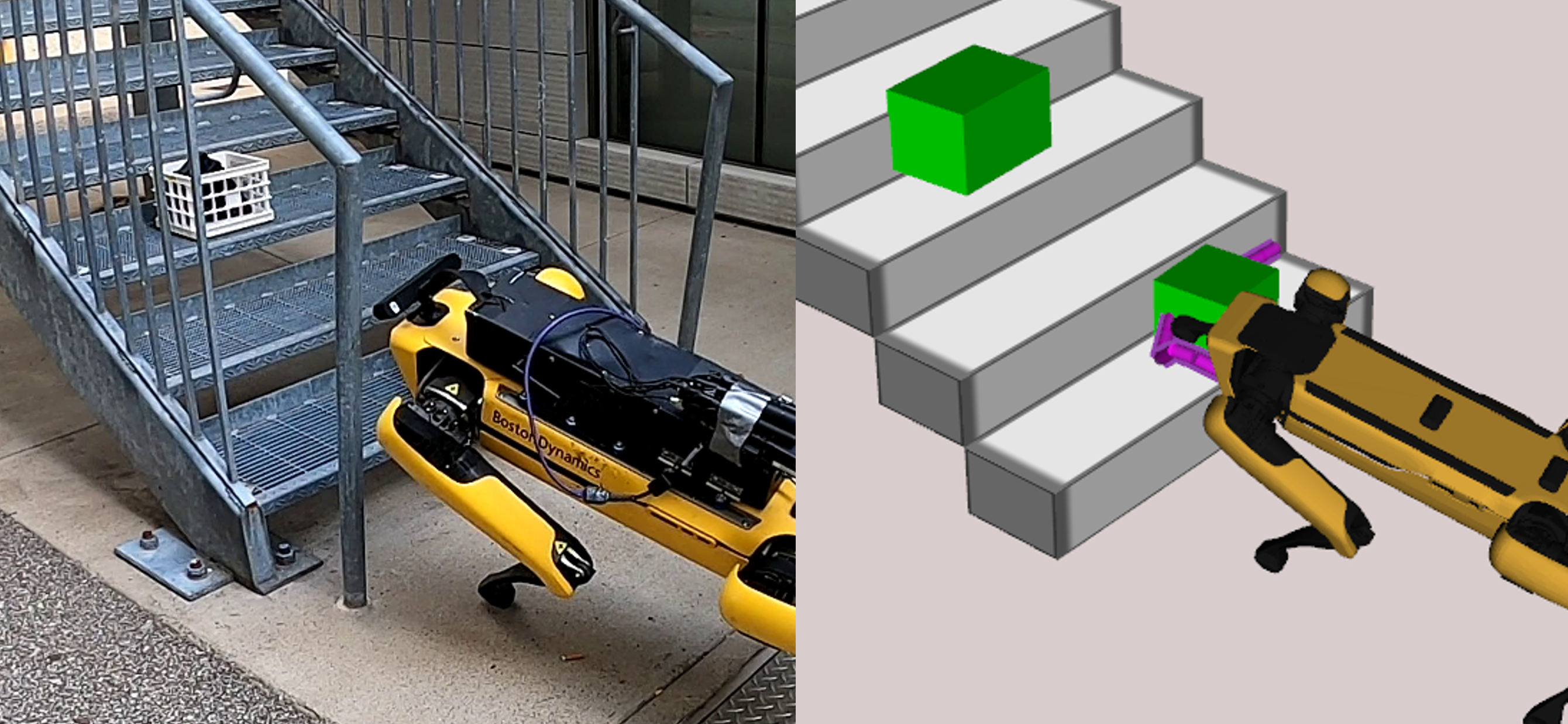}
    \end{subfigure}
    \begin{subfigure}[t]{.32\linewidth}
        \centering
         \includegraphics[width = \linewidth]{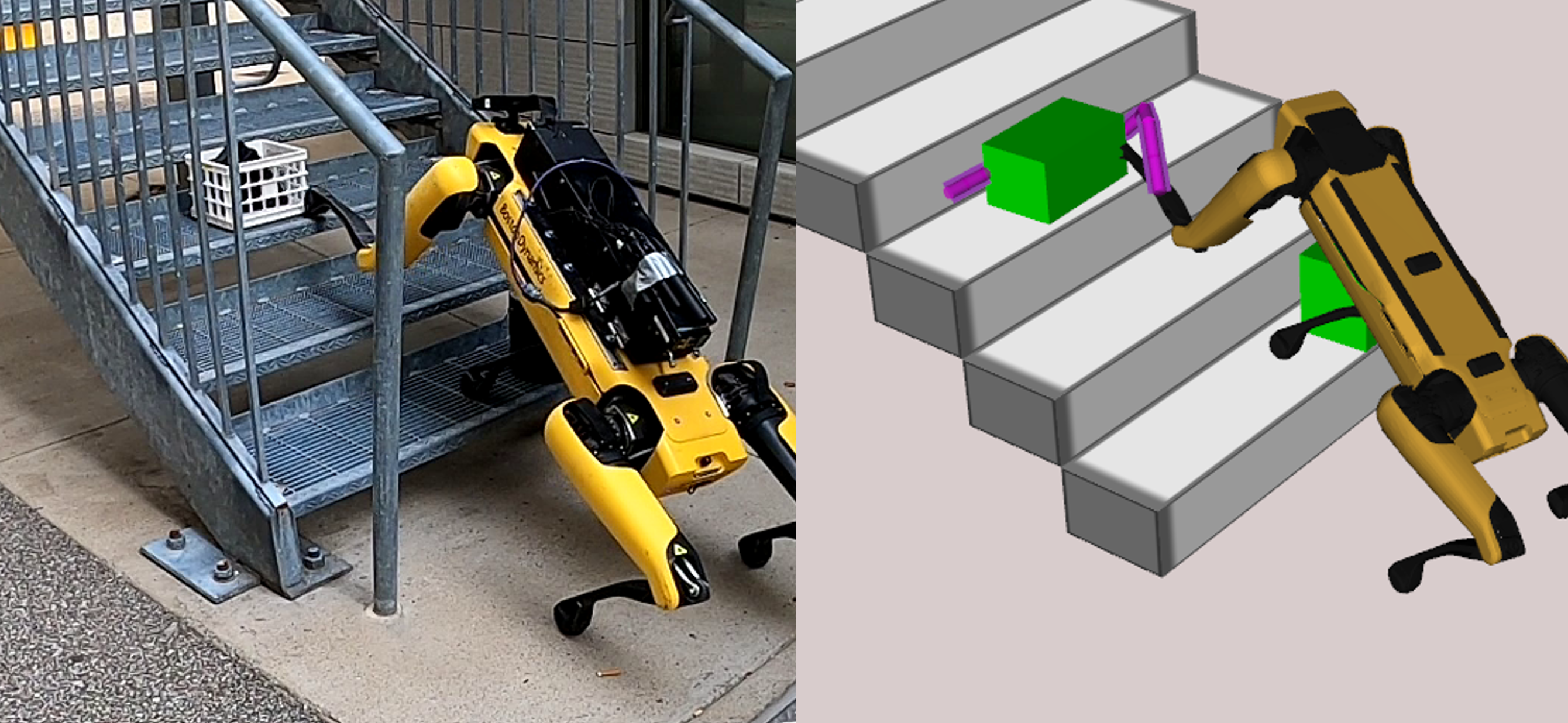}
    \end{subfigure}
    \begin{subfigure}[t]{.32\linewidth}
        \centering
        \includegraphics[width = \linewidth]{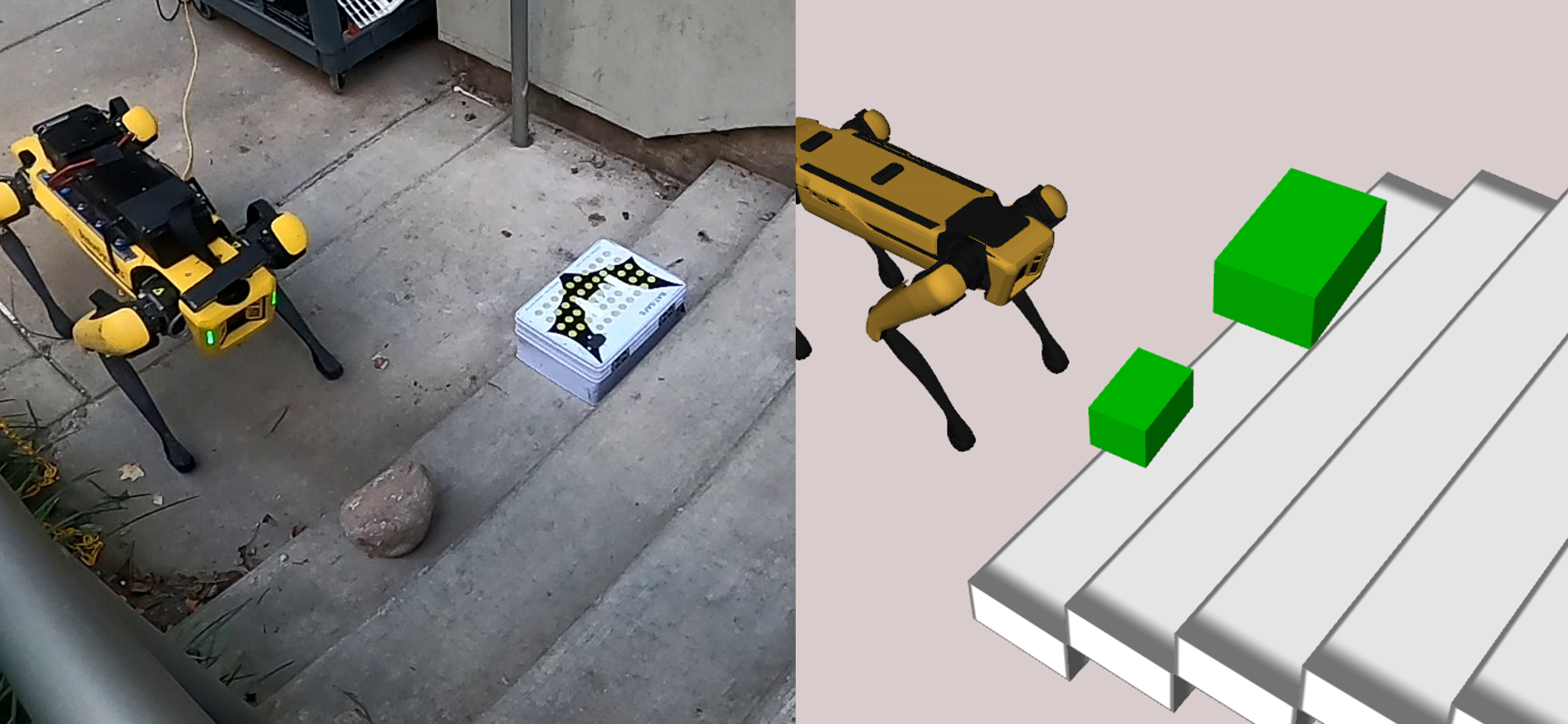}
    \end{subfigure}
    \begin{subfigure}[t]{.32\linewidth}
        \centering
         \includegraphics[width = \linewidth]{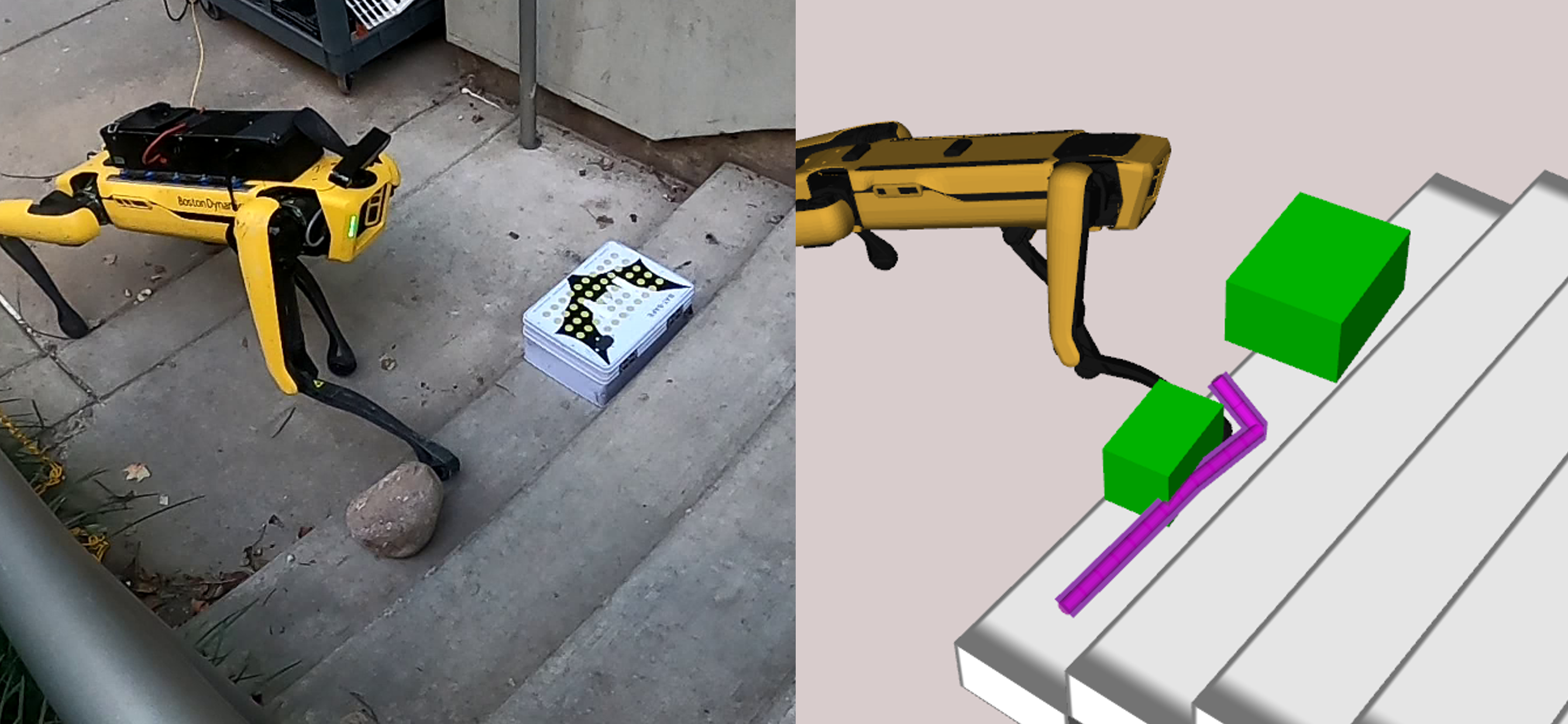}
    \end{subfigure}
    \begin{subfigure}[t]{.32\linewidth}
        \centering
         \includegraphics[width = \linewidth]{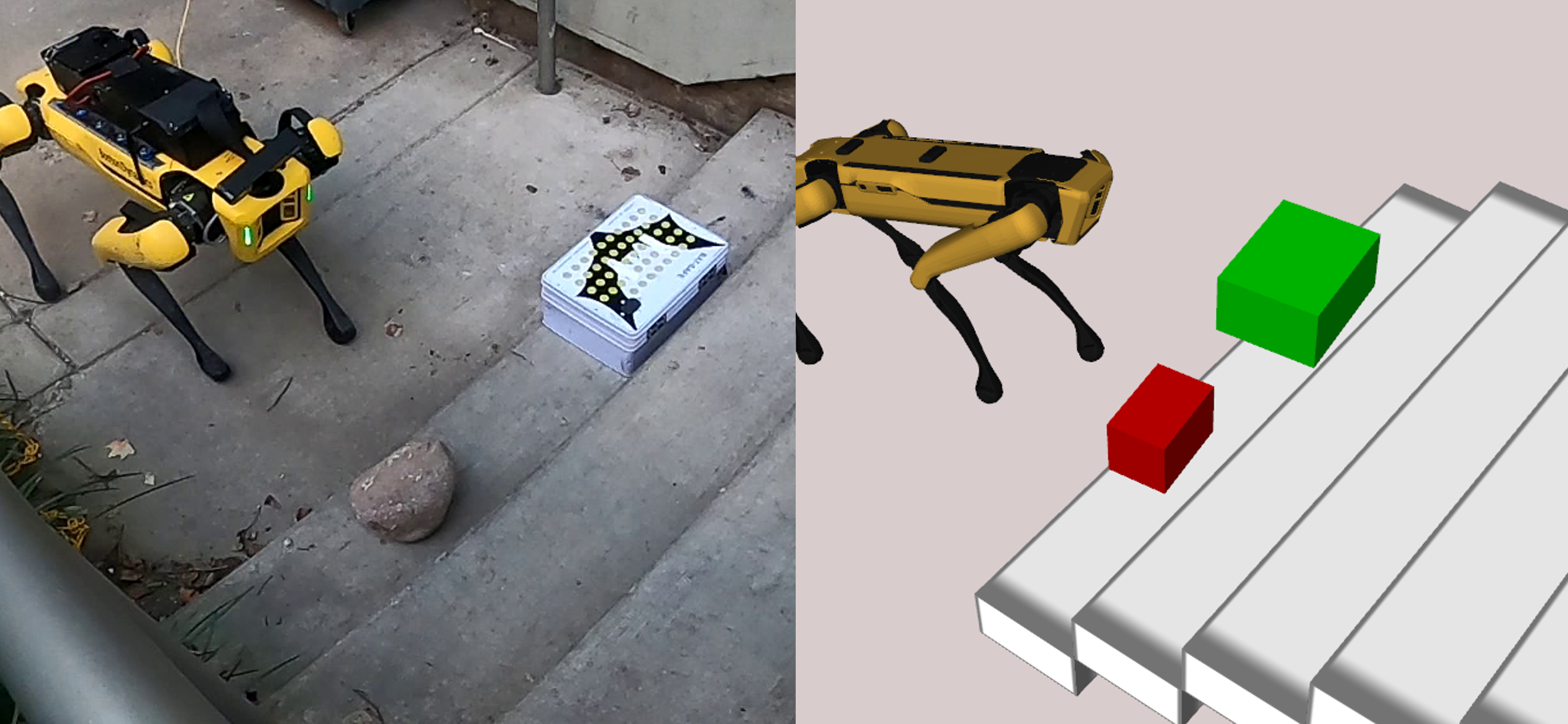}
    \end{subfigure}
    \begin{subfigure}[t]{.32\linewidth}
        \centering
        \includegraphics[width = \linewidth]{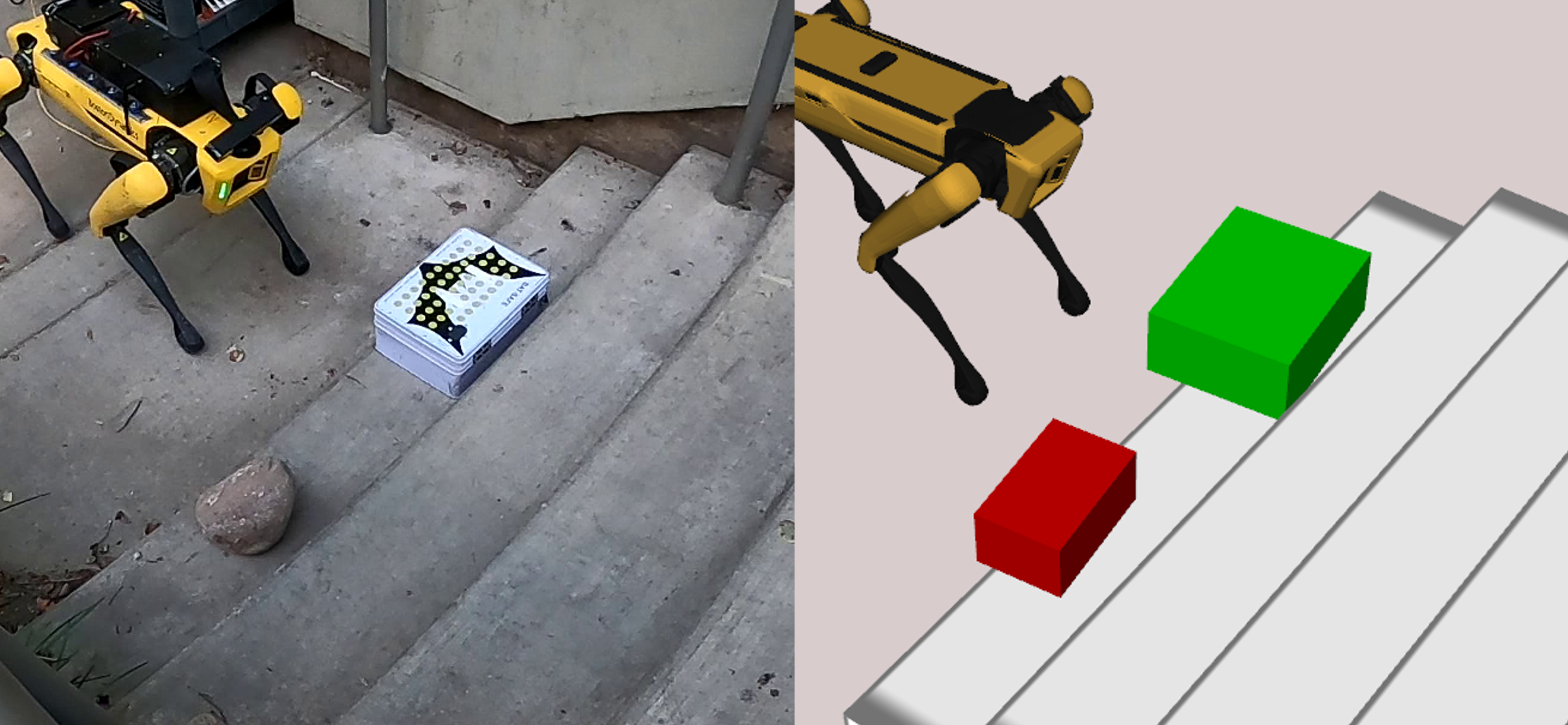}
    \end{subfigure}
    \begin{subfigure}[t]{.32\linewidth}
        \centering
         \includegraphics[width = \linewidth]{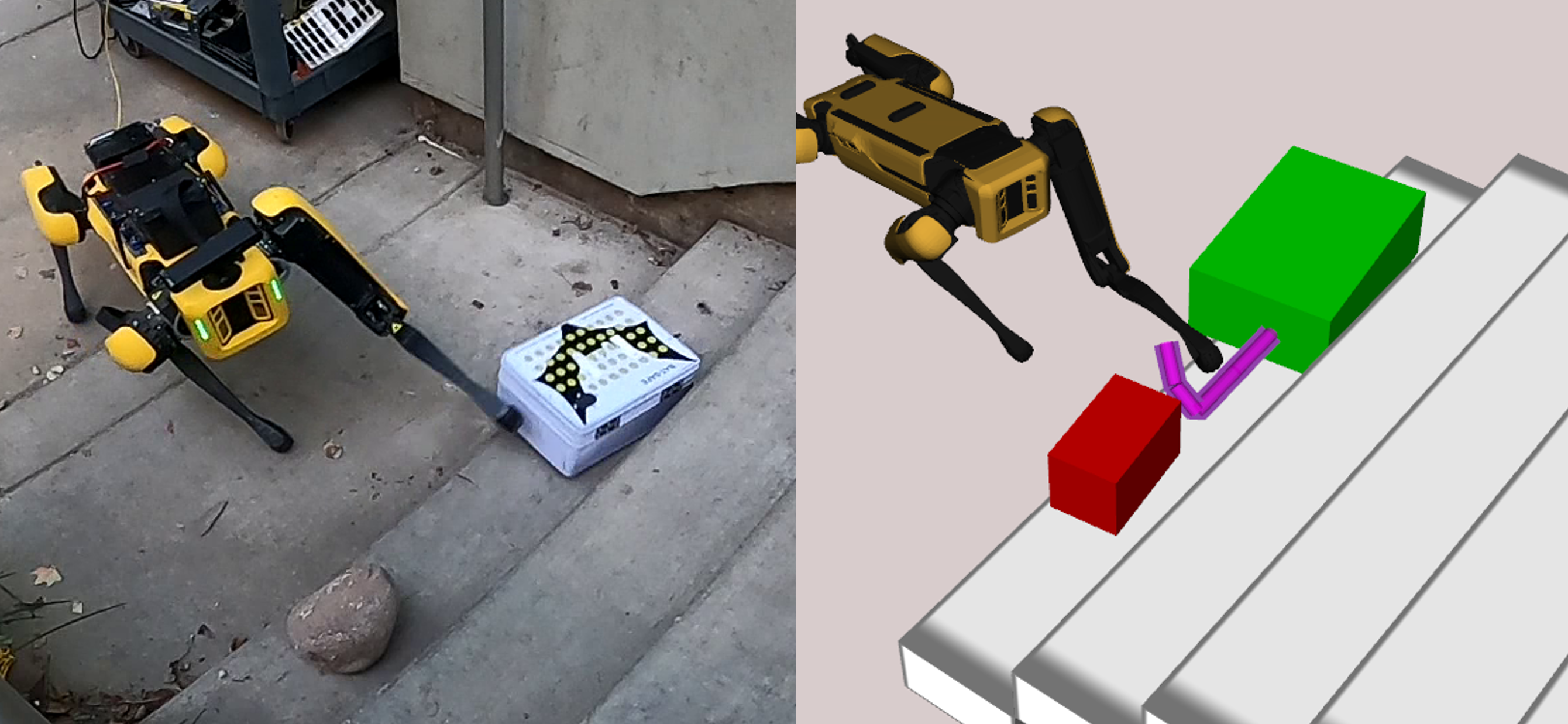}
    \end{subfigure}
    \begin{subfigure}[t]{.32\linewidth}
        \centering
         \includegraphics[width = \linewidth]{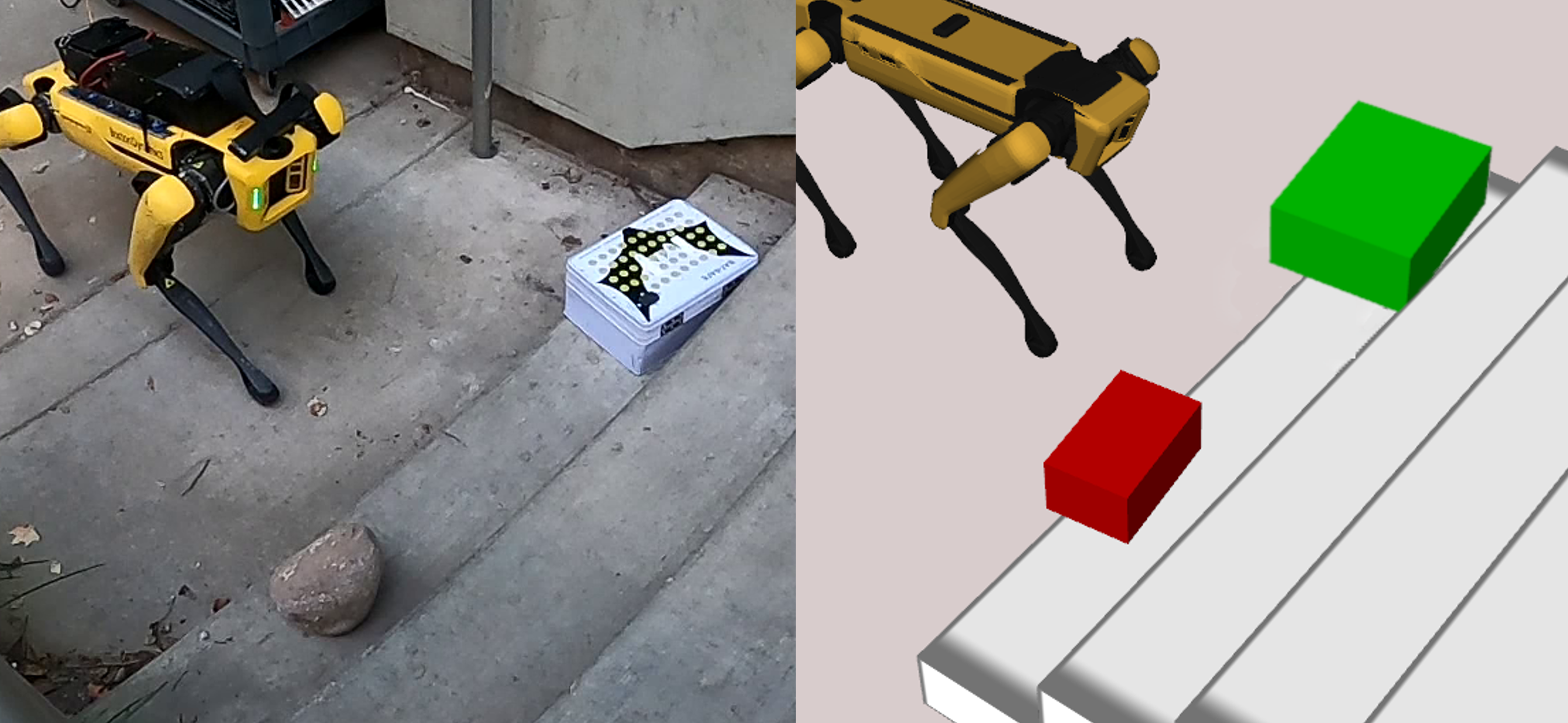}
    \end{subfigure}
    \caption{Boston Dynamics Spot robot executing pushing maneuvers to clear clutter on different types of staircases. Each row shows a push attempt with columns indicating time lapse. The figure shows both real-world experiments (left of each pair) and the corresponding visualization of our perceived world model (right of each pair). In the world model, the staircase is represented by white steps, movable clutter is shown in green (translucent for initial, opaque for predicted/updated position), and objects identified as immovable are red. The planned foot trajectory for the push is shown in magenta. Row 3 demonstrates the system's ability to update movability of heavy objects.}
    \label{fig:results}
    \vspace{-1.5em}
\end{figure*}

Figure \ref{fig:results} illustrates the performance of our complete framework across several real-world trials. Each row demonstrates a distinct scenario, highlighting different capabilities of the system. The top rows show successful tracking of the contact-based predicted location (opaque green) and successful post-interaction visual update, enabling robust re-detection. The bottom two rows demonstrate the system's ability to learn from physical interaction and adapt its plan. In this multi-object scenario, the robot first attempts to push the rock. After making contact and detecting that the push has stalled, it correctly reclassifies the stone as an immovable object in its world model (red). Crucially, the system then proceeds to the next action in its high-level plan, successfully manipulating the second, movable object to clear a path.

Our proposed framework with contact feedback drastically improves task success rate compared to an open-loop baseline, as shown in Fig.~\ref{fig:success_rate}. For instance, our system achieved an 85\% success rate with the paint can, where the baseline struggled at just 35\%, while maintaining a robust 75\% success rate on the challenging metal frame, where the baseline succeeded only 25\% of the time. The metal frame proved particularly challenging due to its lower weight, which made it more difficult for our contact estimation to precisely predict the moment of impact. This enhanced task performance is a direct result of the superior accuracy of our state estimation pipeline (Table~\ref{tab:tracking_comparison}). The mean prediction error for the weighted crate was reduced by an order of magnitude, and the error for the box was reduced by nearly 65\%. Even in this difficult case of the metal frame, our system performed 35\% better than the open-loop baseline. This tracking precision, which also demonstrates significantly lower variance, is the key factor that enables more informed re-planning and transforms unreliable open-loop attempts into consistent successes. These results clearly show that incorporating contact feedback is crucial for reliably handling the uncertainties inherent in the interaction task.

\begin{table}[b!]
  \vspace{-1.5em}
  \centering
  \sisetup{separate-uncertainty = true}
  \resizebox{\columnwidth}{!}{%
    \begin{tabular}{
      l
      S[table-format=1.2(2)]
      S[table-format=1.2(2)]
    }
    \toprule
    & \multicolumn{2}{c}{Mean Prediction Error (m)} \\
    \cmidrule(lr){2-3}
    \textbf{Object} & {\textbf{Ours (with contact feedback)}} & {\textbf{Baseline (open-loop)}} \\
    \midrule
    Box            & $\boldsymbol{ 0.09 \pm 0.04 }$ & 0.26 \pm 0.33 \\
    Paint can      & $\boldsymbol{ 0.09 \pm 0.03 }$ & 0.19 \pm 0.14 \\
    Metal frame    & $\boldsymbol{ 0.15 \pm 0.13 }$ & 0.23 \pm 0.17 \\
    Weighted crate & $\boldsymbol{ 0.03 \pm 0.02 }$ & 0.33 \pm 0.14 \\
    \bottomrule
    \end{tabular}%
  }
  \caption{Comparison of object-level mean prediction error (mean $\pm$ std) in meters.}
  \label{tab:tracking_comparison}
  \vspace{-0.5em}
\end{table}

\subsection{Limitations}
While our framework significantly outperforms the baseline, it has several limitations that present clear avenues for future work. First, the performance of the RL pedipulation policy is sensitive to state estimation errors. Cumulative odometry drift, common when traversing long staircases, can lead to inaccurate foot placements, and the policy's stability is most challenged when pushing from the top of a staircase, which can cause the robot to entirely miss the object. It can also push an object entirely off an open-sided step and lose track of it. Second, our contact estimation method, which relies on residual motor torques, struggles with very lightweight objects. This was observed with the hollow metal frame, where the minimal interaction forces did not always surpass the detection threshold, preventing proprioceptive tracking from activating and leading to higher errors. Finally, the camera's limited field-of-view (FOV) is a challenge. Objects successfully pushed far to the side can land outside the FOV, preventing a post-interaction visual update and causing the system to enter a futile retry loop. Several of these failure cases are presented in the supplemental video.


%% file: sections/conclusions.tex
\section{Conclusion}
In this work, we presented a complete perception-to-action framework that enables quadrupedal robots to robustly clear clutter from staircases through legged pushing. We showed that by tightly integrating perception, proprioceptive state estimation, and hierarchical planning, our system successfully addresses the challenges of occlusion and partial pushes inherent in real-world physical interaction. Experiments across multiple staircases and diverse objects demonstrated that incorporating contact feedback significantly improves state estimation accuracy and overall task success rates compared to open-loop baselines.

Despite these advances, some limitations remain. Performance can degrade under significant odometry drift or when manipulating lightweight objects that produce weak contact signals. Additionally, the limited field-of-view of the onboard camera hinders re-detection when pushing objects on really wide staircases. These factors underscore the need for broader perception, stronger temporal modeling of contact, and tighter integration of planning with prior knowledge. Furthermore, training the pedipulation policy in noisy, cluttered staircase environments could improve stability, particularly during the challenging transition from pushing back to standing balance on the staircase.

Future work will expand the framework along various directions. First, we plan to integrate a higher-level task planner that reasons over sequences of interactions, enabling the robot to decide when and how to attempt pushes based on prior knowledge of object movability. Second, we aim to incorporate explicit priors into the planning loop, allowing the robot to use past interaction outcomes to guide future decisions and reduce retries. Finally, we will pursue better contact estimation through time-series modeling of interaction forces and uncertainty propagation, enabling more accurate distinction between missed contacts, partial pushes, and immovable objects. Together, these directions will move quadrupedal robots closer to long-horizon autonomy, where locomotion, manipulation, and reasoning are seamlessly combined in densely cluttered environments.